\documentclass[sigconf]{acmart}

\usepackage{amsmath}
\usepackage{amssymb}
\usepackage{amsfonts}
\usepackage{graphicx}
\usepackage{caption}
\usepackage{bm}

\def\BibTeX{{\rm B\kern-.05em{\sc i\kern-.025em b}\kern-.08emT\kern-.1667em\lower.7ex\hbox{E}\kern-.125emX}}
    
\copyrightyear{2019}
\acmYear{2019} 
\setcopyright{iw3c2w3}
\acmConference[WWW '19 Companion]{Companion Proceedings of the 2019 World Wide Web Conference}{May 13--17, 2019}{San Francisco, CA, USA}
\acmBooktitle{Companion Proceedings of the 2019 World Wide Web Conference (WWW '19 Companion), May 13--17, 2019, San Francisco, CA, USA}
\acmPrice{}
\acmDOI{10.1145/3308560.3316607}
\acmISBN{978-1-4503-6675-5/19/05}

\usepackage{balance}

\begin{document}

%
\title{ProductNet: a Collection of High-Quality Datasets for Product Representation Learning}

\author{Chu Wang}
\affiliation{Amazon.com}
\email{chuwang@amazon.com}

\author{Lei Tang}
\affiliation{Amazon.com}
\email{leitang@amazon.com}

\author{Yang Lu}
\affiliation{Amazon.com}
\email{ylumzn@amazon.com}

\author{Shujun Bian}
\affiliation{Amazon.com}
\email{sjbian@amazon.com}

\author{Hirohisa Fujita}
\affiliation{Amazon.com}
\email{hirohisf@amazon.com}

\author{Da Zhang}
\affiliation{Amazon.com}
\email{dazh@amazon.com}

\author{Zuohua Zhang}
\affiliation{Amazon.com}
\email{zhzhang@amazon.com}

\author{Yongning Wu}
\affiliation{Amazon.com}
\email{yongning@amazon.com}

%
\renewcommand{\shortauthors}{Chu Wang et al.}

%
\begin{abstract}
ProductNet is a collection of high-quality product datasets for better product understanding. Motivated by ImageNet, ProductNet aims at supporting product representation learning by curating product datasets of high quality with properly chosen taxonomy. In this paper, the two goals of building high-quality product datasets and learning product representation support each other in an iterative fashion: the product embedding is obtained via a multi-modal deep neural network (master model) designed to leverage product image and catalog information; and in return, the embedding is utilized via active learning (local model) to vastly accelerate the annotation process. For the labeled data, the proposed master model yields high categorization accuracy ($94.7\%$ top-1 accuracy for 1240 classes), which can be used as search indices, partition keys, and input features for machine learning models. The product embedding, as well as the fined-tuned master model for a specific business task, can also be used for various transfer learning tasks.
\end{abstract}

%
%



%
\maketitle


\section{Introduction}

E-Commerce retail is all about products, be it physical goods or digital content. Product representation learning is the key to enhancing search and discovery experiences for customers, and product management for the backend systems. Compared to raw attributes like image and catalog information, the product representation has advantages of maintaining product semantic information and being in a compact form to facilitate modeling and algorithm designs. Therefore, the major problem becomes how to build high-quality product representation with satisfactory transferability. In order to learn how customers understand products, motivated by the success of ImageNet, we took a data first and quality first approach to facilitate product representation learning, by developing ProductNet, a collection of high-quality labeled product datasets. 

Building high-quality datasets and categorizing products into thousands of classes are difficult; developing product embedding with satisfactory transferability to incorporate product image and text information is even more challenging. To achieve these two goals, instead of utilizing large-volume noisy data, we choose to use small-volume, high-quality {\it gold datasets}. That being said, we explore the quality dimension of the data before the volume dimension by working on a classification task via human annotation. The two goals of product categorization and representation learning support each other in an iterative fashion within our work: the product embedding is built based on deep classifiers for the labeled dataset; the embedding itself, in return, is utilized in an active learning framework to enhance the labeling speed and quality.

Note that the goal of annotation is not to cover billions of products. Instead, we focus on a subset of high-quality products for the representation learning purpose. In particular, we aim at the diversity and representativeness of the products. Being representative, the labeled data can be used as reference products to power product search, pricing, and other business applications. Being diverse, the models are able to achieve strong generalization ability for unlabeled data, and the product embedding is also able to represent richer information. A carefully chosen taxonomy is important to reduce annotation ambiguities and mistakes. We adopt a function-based product taxonomy for the ProductNet construction. As for now, we have populated 3900 categories of non-media products, with roughly 40-60 products for each category. For either categorization or representation learning, we need to handle products with noisy attributes, missing fields, or even manually altered product information. Combining different attributes of products is one way to alleviate these potential problems and to achieve better robustness. We utilize multi-modal learning to incorporate information from different fields into the feature embeddings and to deal with the issue of noisy or missing product attributes.

We are able to achieve $94.7\%$ top-1 accuracy on our product categorization task (1240 classes). The high-accuracy comes from both the multi-modal model and the curated dataset with carefully chosen taxonomy. Such high accuracy opens the door for the predicted categorization to be used as search indices and partition keys. We hope carefully curated product datasets, like ProductNet, will lead a better way for product categorization and product representation learning. Before going into details, we first discuss the challenges of building high-quality datasets and related works.

{\it Challenges.}
Building a high quality dataset is never a trivial task and it faces several challenges: \romannumeral1) the annotation quality to provide correct labels; \romannumeral2) the annotation quantity to support a large scale dataset; \romannumeral3) noisy or missing fields (e.g. misplaced product description or empty product image); \romannumeral4) distinct data format from different fields like categorical attribute, free-form text, and images. 

{\it Related work.}
ImageNet Large Scale Visual Recognition Challenge (ILSVRC), which collects more than 1.2 million varied images, has inspired the development of a series of deep learning models~\cite{krizhevsky2012imagenet,simonyan2014very,szegedy2017inception,he2016deep}.The last hidden layer signals of deep models are widely used as image feature embeddings for transfer learning~\cite{girshick2014rich}; the upper-layers of deep models can also be fine-tuned for different tasks. Natural language processing benefits largely from the introduction of word2vec~\cite{mikolov2013distributed} and its extension to sentences, documents, and sub-word information~\cite{le2014distributed,devlin2018bert}. In addition to ImageNet, there are various datasets which contribute significantly to model development and real-world applications~\cite{mcauley2013hidden,lin2014microsoft}.

\section{Dataset Construction\label{sec:construction}}

In this section, we demonstrate and discuss how we construct ProductNet datasets. 
A naive way for obtaining product labels is directly via human annotation:
given a {\it candidate pool}, each product is tagged with a category name by human.
This way of labeling is unbearably expensive as long as the number of categories is moderately large.
Furthermore, since the annotation involves choosing the correct category out of many, 
background knowledge is required for the labeler and mistakes can not be avoided.

Even though human annotation itself is irreplaceable, we proceed in an intelligent way in order for higher efficiency and data quality.
The entire workflow of ProductNet dataset construction is based on an iterative loop of human annotation and representation learning. 
Before going into a detailed demonstration, we would like to highlight the importance of taxonomy in the construction of labeled datasets.
Product understanding and representation learning can largely benefit from a taxonomy that approximates the intrinsic data distribution well.
There are different types of taxonomies, including function-based, subject-based, and organization-based ones.
To avoid ambiguity and improve product embedding, we prefer the function-based taxonomy for non-media products.

\subsection{Human annotation and local model}
We annotate products category by category.
In this way, an annotator only needs to make a binary decision on whether a product belongs to the current category (positive) or not (negative).
In the meantime, only the background knowledge for the current category is required.
We maintain a set of {\it local models} to provide suggestions of products to be annotated based on available labels.
The local model can be a KNN search engine or a keyword-based search engine.
By searching similar products from the pool, the engine is able to provide relevant candidate products.

\begin{figure*}[hbt!]
\centering
\includegraphics[width=0.55\textwidth]{{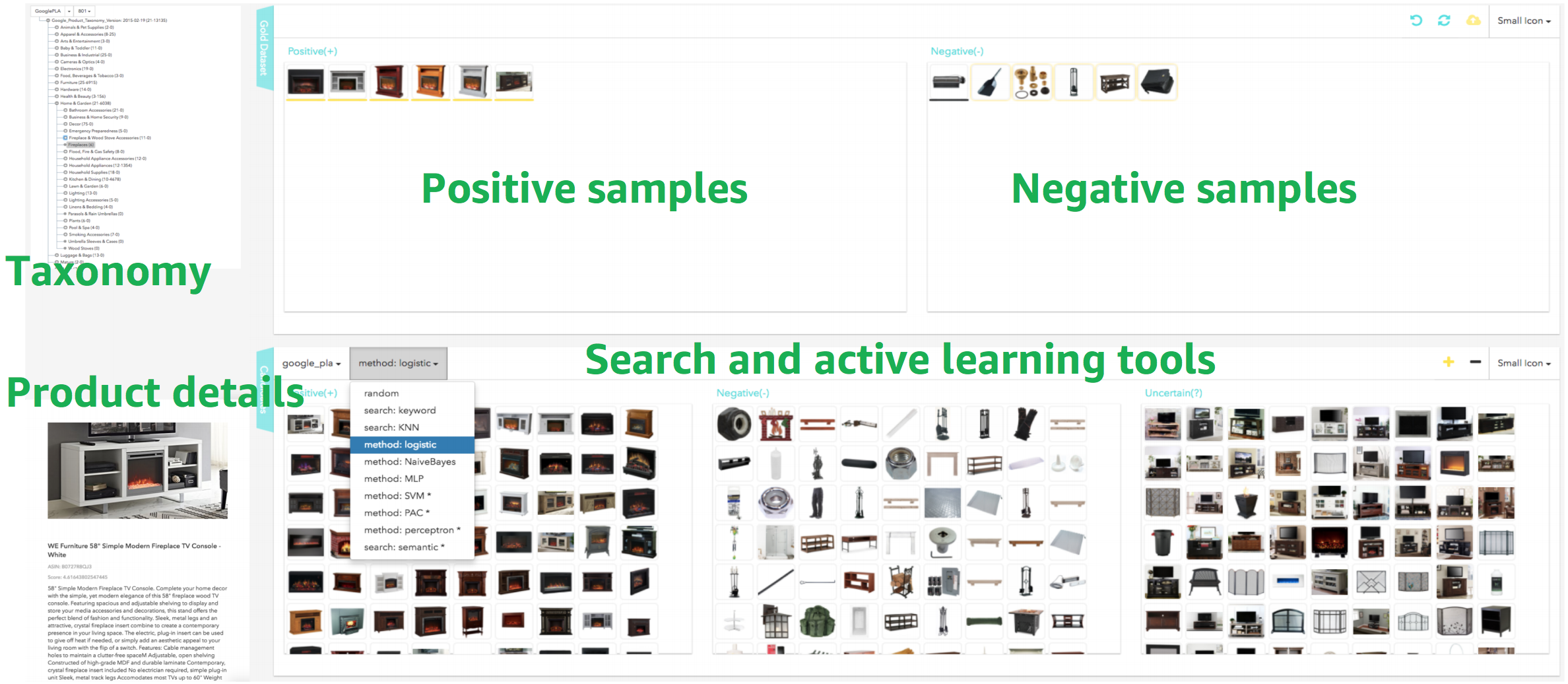}}
\caption{Screenshot of ProductNet annotation portal.\label{fig:portal}}
\end{figure*}

Binary classifiers based on active learning mechanism are more helpful as local models.
The binary classifiers are built on and refined by both the positive and negative samples.
With at least one positive product and one negative product, a binary classifier is trained to initialize the active learning process,
which is then able to provide positive and negative suggestions for labeling.
More importantly, the algorithm asks for labels of ambiguous candidates in order to improve the local model itself,
thus the quality of positive and negative suggestions becomes better.
To promote better variety and retrievability of the products, we use KNN search, keyword search, and active learning methods on generalized linear models, in a mixed way.
A screenshot of ProductNet annotation portal is demonstrated in Figure \ref{fig:portal},
and more discussion of the active learning sampling is in Section \ref{sec:active}

\subsection{Representation learning and master model}

When we have labeled positive data from multiple categories, a multi-modal classification neural network is trained for the goal of representation learning.
We name this deep neural network the {\it master model}.
Based on the gold dataset, we train the master model and extract the last hidden layer signals as product feature embeddings.
The embeddings are then fed back to the active learning module for another round of annotation. 
In addition to the feature embeddings produced by the master model, we also get the product categorization.
As long as the master model yields high accuracy, its prediction also provides accurate product candidate for annotation.

\begin{figure*}
\centering
\includegraphics[width=0.55\textwidth]{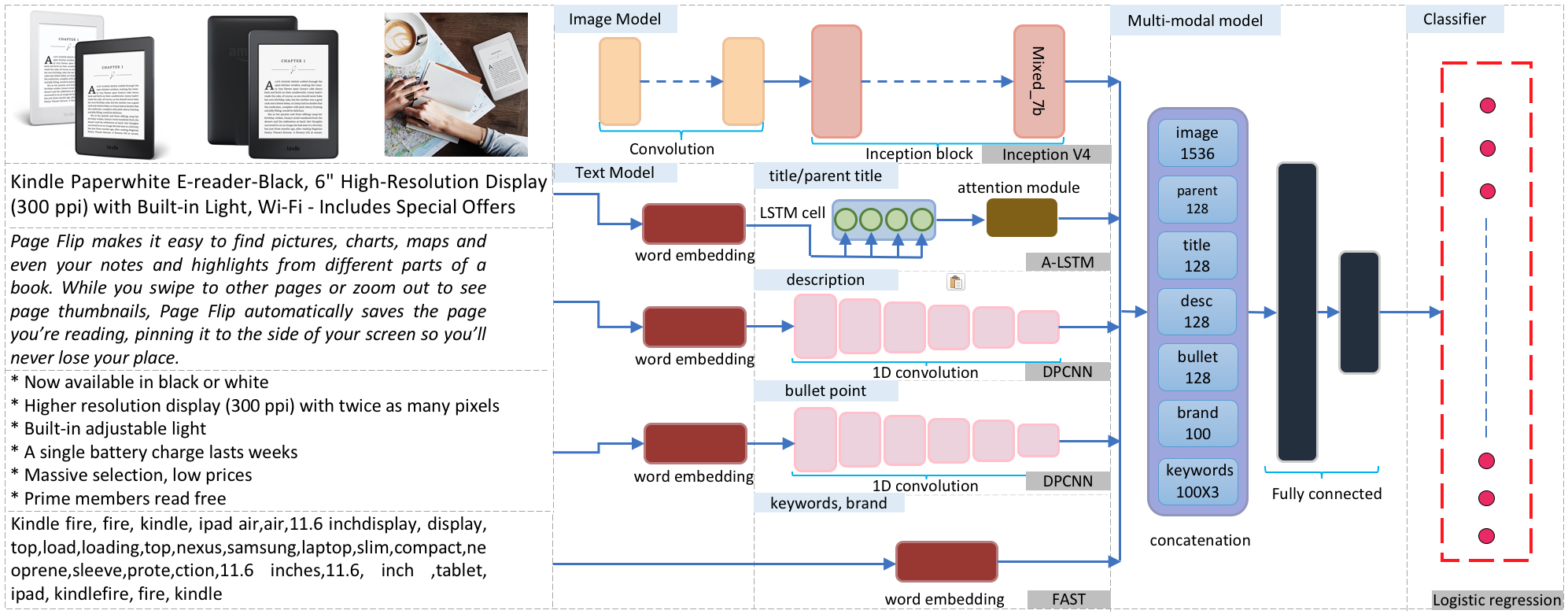}
\caption{Master model structure: multi-modal deep neural network.\label{fig:master}}
\end{figure*}

Formally, let $c=(1, \cdots, C)$ be the class label of a product, $x$ be the input data representing both text and image data sources.
We learn a conditional distribution $p(c|x)$ defined as
\begin{equation}
 p(c|x) = \frac{ \exp\{f_c(x; w) +b_c\} }{\sum_{c=1}^{C} \exp \{ f_c(x;w) +b_c\}}
\end{equation}
where $f_c(x;w)$ is the scoring function from class $c$, $b_c$ is the bias term, and $w$ collects all trainable weights. The scoring function $f_c(x;w)$ fuses both text and image information computed by neural networks.
The complete structure of our master model is shown in Figure \ref{fig:master}.
We discuss our design of the master model from different aspects in the following several sessions.

We adopt Inception-v4 as our image model~\cite{szegedy2017inception}. When the image is missing for a product, a zero image is used instead. The output of image model is a 1536-dimensional vector from mixed\_7d layer from Inception-v4 for model fusion.
There are seven fields of catalog data that we have considered: product title, product description, bullet points, brand, and three types of keywords. In spite of missing fields and varying qualities, the multi-modal deep neural network is able to adjust the relative weights for better results.
We choose deep pyramid convolutional neural network (DPCNN) for description and bullet points, attention-LSTM (A-LSTM) for product title, and vanilla fasText (FAST) for the brand and keywords~\cite{johnson2017deep,vaswani2017attention,sak2014long,joulin2016bag}.
The models for each field are selected by running experiments on each field alone and adopting the one with the best performance.
A multi-modal model builds a joint representation of different data sources and they could reinforce each other. 
We learn a fused model integrating both text signals and image signals. 
The final feature vector is composed by concatenating the outputs from image and text models with two fully-connected layers.
Note that the dimension of the feature embeddings is adjustable, which is
helpful when low-dimension embedding is needed for higher efficiency.

\subsection{Iterative construction of datasets}
The representation learning module and the human annotation module operate iteratively.
For any given candidate pool and taxonomy, initial product embeddings are required to start the human annotation process.
The initial embeddings can be obtained via pre-trained models.
For example, we choose Inception-v4 trained on ImageNet for image data, and fastText model for text data.

Note that it is not necessary to use high quality embeddings initially.
The initial feature embeddings will support retrieval methods like active learning for the first round of annotation, and then the labels are used for training the master model and producing a new version of product embeddings. With the new embeddings, we conduct another human annotation stage. But this time, the speed and quality are further boosted by the embeddings. 

One of the advantages of our iterative construction is the ability of local adjustment.
Note that if we want to apply active learning directly to the master model, the newly labeled data will have very small impact on the model because the number of classes is large. With the local models, on the other hand, we are able to conduct local adjustment precisely and only focus on the categories of interest.
Furthermore,
we frequently mark the wrongly labeled data by the master model as local negatives, so that further annotated data will direct the master model to the more accurate direction.


\section{Experiments\label{sec:experiment}}

In this section, we demonstrate our experiment results. As a proof of concept, we adopt the Google Taxonomy~\cite{googletax}, which partitions products based on their functionalities in a tree structure. As the paper submitted, our labeled dataset covers 3900 leaf nodes with 178k products. Most of products have both text and image information. Each leaf node is labeled by one labeler and verified by one auditor independently to reduce labeling errors. At each leaf node, we also maintain a similar number of negative samples which are most likely to be confused with positive ones for labeling reference and training the local model. 
To evaluate our dataset, we first exhibit some qualitative results regarding the acceleration of labeling by active learning,
then evaluate the accuracy of mater model prediction, and discuss how combining the two accelerates the dataset construction in the end.

\subsection{Active learning recommendation\label{sec:active}}
\begin{figure}
\centering
\includegraphics[width=0.45\textwidth]{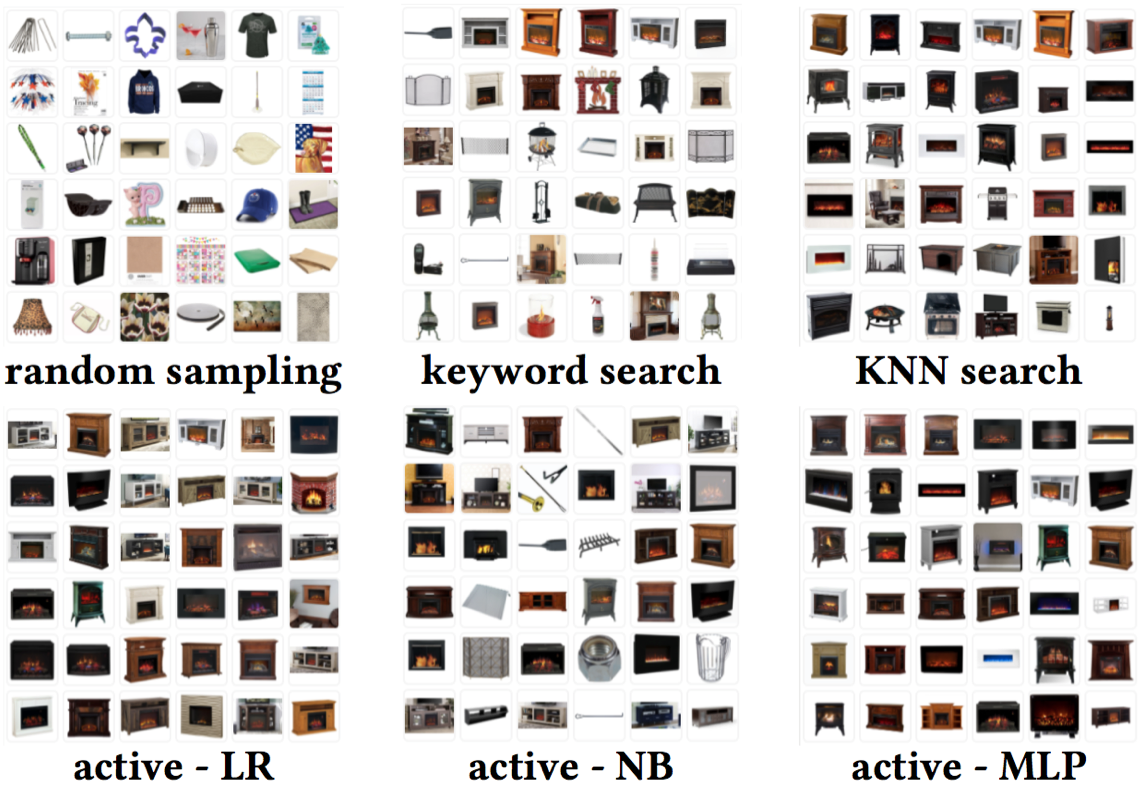}
\caption{Comparison of different sampling methods.\label{fig:local}}
\end{figure}

We demonstrate how our local model and active learning technique accelerate the labeling process.
The product retrieval module via active learning and other methods is located in the right bottom of Figure \ref{fig:portal},
Currently, we support random sampling, keyword search, KNN search, ad-hoc input, and active learning based on logistic regression (LR), naive Bayes (NB), or multi-layer perceptron (MLP).
As an example, we demonstrate the sampled products for the leaf node ``Fireplaces'' in Figure \ref{fig:local}.
It can be observed that random sampling can hardly yield any fireplace from the candidate pool;
the keyword search is able to get several fireplaces while most of the search results are irrelevant;
active learning and KNN search are able to provide meaningful recommendations to enhance the labeling speed by a significant amount.
Furthermore, the active learning methods and KNN search are able to provide different types of fireplaces to further enhance product variety.

\subsection{Master model classification}
We test the master model on a subset of 1240 leaf nodes, each having 40 products.
For each category, 32 products (80\%) are randomly chosen as the training set, and the rest as test set.
The classification accuracy is high, indicating the powerful generalization ability of our multi-modal model.
The quality of our dataset also contributes to the high accuracy, as any defects within categories will lower the prediction accuracy.
The training process takes 10 hours on a single AWS p3.2xlarge GPU instance for 50 epochs to achieve 94.7\% accuracy and become stabilized.
We demonstrate the performance of the master model with image and text sources (master-IT) and show its improvement with non-multimodal methods like Inception-v4, bag-of-words (BOW), and text-only master model (master-T) in Table \ref{table:master}.
Though the high classification accuracy demonstrates the advantage of the master model structure, we would like to highlight the contribution of the dataset being of high quality. With a properly designed taxonomy and accurately annotated products, the demand for sophisticated models decreases, while the model transferability increases.
\begin{table}[htp]
\caption{Master model classification accuracy (percentile).\label{table:master}}
\centering
\begin{tabular}{crrrr} 
\toprule
			&	Inception-v4	&	 BOW & master-T		&	master-IT		\\
\midrule
top-1	 acc	&	69.4   		&		83.1	&	92.6   	&		94.7 		   \\
top-3	 acc	&	85.5   		&		90.7	&	97.8   	&		98.2   	  \\
top-5	 acc	&	90.3   		&		93.1	&	98.5  	 &		99.1  		  \\
\bottomrule
\end{tabular}
\end{table}
\vspace*{-\baselineskip}

\subsection{Annotation Acceleration}
We test the acceleration of the active learning retrieval for the annotation, and find that a normal annotator with general background knowledge is able to label 100 positive data points for each leaf node within 30 minutes. Compared to vanilla labeling which takes 30 minutes to find 5 positive data points for a leaf node, the acceleration factor is roughly 20.
Furthermore, after the master model is trained, the products from the annotation pool will be given a recommended category by the master model so that the annotation process can be further accelerated by verifying that category label, and we estimate the annotation acceleration factor via master model is 80 compared to vanilla labeling.
 

\section{Conclusion\label{sec:conclusion}}

We have introduced ProductNet, a collection of high-quality product datasets for better product understanding.
Our framework is a fast and reliable way of constructing product labels and building high-quality datasets.
The master model is able to provide business acceptable labels for product listings, product indexing, and partition keys;
and the product embedding obtained can support various product modeling tasks and business applications.
The experiments verify our initiative that a dataset of high-quality is able to foster high-quality product embeddings.

%

%
\bibliographystyle{ACM-Reference-Format}
\bibliography{productnet}

%

\end{document}